\pdfoutput=1

\documentclass[11pt]{article}

\usepackage[preprint]{acl}

\usepackage{times}
\usepackage{latexsym}

\usepackage[T1]{fontenc}

\usepackage[utf8]{inputenc}

\usepackage{microtype}

\usepackage{inconsolata}

\usepackage{graphicx}

\hypersetup{breaklinks=true} 

\title{A Japanese-Chinese Parallel Corpus Using Crowdsourcing \\
  for Web Mining}

\author{Masaaki Nagata, Makoto Morishita\thanks{Currently at
    Future Corporation}, Katsuki Chousa, Norihito Yasuda\\
  NTT Communication Science Laboratories, NTT Corporation \\
  2-4 Hikaridai Seika-cho Souraku-gun Kyoto 619-0237 Japan \\
  \texttt{\{masaaki.nagata,katsuki.chousa,norihito.yasuda\}@ntt.com}
  \\}

\begin{document}
\maketitle
\begin{abstract}
  Using crowdsourcing, we collected more than 10,000 URL pairs
  (parallel top page pairs) of bilingual websites that contain
  parallel documents and created a Japanese-Chinese parallel corpus of
  4.6M sentence pairs from these websites.  We used a Japanese-Chinese
  bilingual dictionary of 160K word pairs for document and sentence
  alignment.  We then used high-quality 1.2M Japanese-Chinese sentence
  pairs to train a parallel corpus filter based on statistical
  language models and word translation probabilities.  We compared the
  translation accuracy of the model trained on these 4.6M sentence
  pairs with that of the model trained on Japanese-Chinese sentence
  pairs from CCMatrix (12.4M) \cite{schwenk-etal-2021-ccmatrix}, a
  parallel corpus from global web mining.  Although our corpus is only
  one-third the size of CCMatrix, we found that the accuracy of the
  two models was comparable and confirmed that it is feasible to use
  crowdsourcing for web mining of parallel data.\footnote{Work in
    progress}
\end{abstract}

\section{Introduction}

Parallel data is vital in machine translation for traditional
encoder-decoders and recent large language models. From an analysis of
Palm's training data, \citet{briakou-etal-2023-searching} showed that
large language models could translate because their training data
contain parallel data incidentally.

Relatively small large language models of around 10B parameters have
poor translation accuracy.  However, \citet{Xu_etal_ICLR2024} proposed
ALMA, a method of fine-tuning an LLM on a large amount of monolingual
data, followed by fine-tuning on a small amount of high-quality
bilingual data. They achieved translation accuracy comparable to GPT-3
using relatively small LLMs. \citet{Guo_etal_arXiv2024} showed that
continuous pre-training of large language models on large amounts of
parallel data before fine-tuning on high-quality bilingual data
improves translation accuracy over ALMA.


This paper discusses a method for collecting Japanese and Chinese
parallel sentence pairs from the web. Translation between Japanese and
Chinese is considered one of the most important non-English language
pairs in terms of the number of speakers and the scale of the economy.
We specifically report on the effectiveness of crowdsourcing in
collecting URLs of websites containing parallel data.

\citet{morishita-etal-2022-domain} proposed a method of collecting
parallel URLs (parallel page pairs) using cloud workers for domain
adaptation of machine translation.  We collected parallel top page URL
pairs of bilingual websites by specifying only language pairs to the
crowd workers, with no particular restriction on the target domain.

The experiment results show crowdsourced websites can collect more
parallel sentence pairs with less crawling than automatically
collected bilingual websites using Common Crawl.  We also show that
the translation accuracy achieved using the Japanese-Chinese parallel
corpus created using crowdsourcing is comparable to that achieved
using Japanese-Chinese pairs of CCMatrix
\cite{schwenk-etal-2021-ccmatrix}, a parallel corpus created by global
web mining.  It is worth noting that our corpus only contains
one-third of the data present in CCMatrix.

We release a 4.6M Japanese-Chinese parallel corpus created using
crowdsourcing as JParaCrawl Chinese v2.0 for research purposes only.
\footnote{\url{https://www.kecl.ntt.co.jp/icl/lirg/jparacrawl/}}

\section{Related Work}

\subsection{Web mining for parallel data}

Research on collecting bilingual data by mining the web began around
2000 \cite{resnik-1999-mining,uszkoreit-etal-2010-large}.  We can
broadly divide current research into hierarchical mining (local
mining) and global mining.

In hierarchical mining (or local mining), based on the web's
hierarchical structure, we first search for websites that include
parallel documents, then search for parallel document pairs within a
website, and then search for parallel sentence pairs within a parallel
document pairs. In global mining, we consider the web a flat, massive
set of sentences.  We use similarity based on a multilingual sentence
embedding model to find a sentence's translation among all sentences
in the web in different languages.  ParaCrawl
\cite{banon-etal-2020-paracrawl} is a prime example of the former, and
CCMatrix \cite{schwenk-etal-2021-ccmatrix} is a prime example of the
latter.

Most previous parallel corpora are created by local mining.  Like
EuroParl \cite{koehn-2005-europarl} and OpenSubtitles
\cite{lison-tiedemann-2016-opensubtitles2016}, we first identify a
website that contains parallel documents, extract bilingual document
pairs based primarily on metadata, and then perform sentence
alignment.

The first successful example of global mining is WikiMatrix
\cite{schwenk-etal-2021-wikimatrix}, which collected bilingual text
pairs from Wikipedia in various languages using
LASER\footnote{\url{https://github.com/facebookresearch/LASER}}, a
multilingual sentence embedding model.  CCMatrix
\cite{schwenk-etal-2021-ccmatrix} applies global mining to CCNet
\cite{wenzek-etal-2020-ccnet}, a monolingual corpus of various
languages extracted from Common Crawl.  In the No Language Left Behind
(NLLB) project \cite{nllb2022}, they extended CCMatrix to over 200
languages.

Performing global mining for the entire web requires enormous
computational resources. We consider local mining a more realistic
approach to collecting bilingual data for specific language pairs, as
in the case of JParaCrawl
\cite{morishita-etal-2020-jparacrawl,morishita-etal-2022-jparacrawl},
which collected parallel data between Japanese and English.

In local mining, many previous works on document and sentence
alignment exist, but little research has been done on how to find
websites that contain parallel data.  ParaCrawl
\cite{banon-etal-2020-paracrawl} used the Common Crawl archives to
collect websites that contain bilingual text. They applied a language
detector on each site's web pages and looked for sites that contained
approximately the same amount of text in the language pair to be
collected.  CCAligned \cite{el-kishky-etal-2020-ccaligned} analyzed
the Common Crawl archives using language-identifiable strings in URLs
as clues and collected parallel URL pairs.
\citet{morishita-etal-2022-domain} used a cloud worker to collect
parallel URLs for domain adaptation of machine translation.


\subsection{Japanese-Chinese parallel corpora}

The most widely used Japanese-Chinese parallel corpus for research is
Japanese-Chinese ASPEC (Asian Scientific Paper Excerpt Corpus), which
has 0.68 million sentence pairs consisting of abstracts of Japanese
scientific papers and their manual translation into Chinese
\cite{nakazawa-etal-2016-aspec}.  The JPO-NICT Chinese-Japanese
parallel
corpus\footnote{\url{https://alaginrc.nict.go.jp/jpo-outline.html}},
which has about 130 million Japanese-Chinese patent sentence pairs,
extracted from patent applications in Japan and China based on patent
families.  ASPEC and JPO-NICT are parallel corpora for specific fields
and unsuitable for general translation between Japanese and Chinese.




WCC-JC 3.0 \cite{Zhang_etal_ApplSci2022,Zhang_etal_Electronics2023} is
a Japanese-Chinese parallel corpus of approximately 3M sentence pairs
collected from the web, including movie and TV subtitles, lyrics, and
news articles.
It is available for research purposes by sending an email request to
the authors.

Japanese-Chinese bilingual data collected from Wikipedia include
LinguaTools-WikiTitles v2014 (1.7M sentence pairs), which is contained
in OPUS \cite{tiedemann-2012-parallel}, a collection of open parallel
corpora, WikiMatrix (1.3M sentence pairs)
\cite{schwenk-etal-2021-wikimatrix}, and Wikipedia Chinese-Japanese
Parallel Corpus (0.13M sentence pairs)
\cite{chu-etal-2014-constructing,Chu_etal_TALLIP2015}.  OpenSubtitles
v2018
\cite{lison-tiedemann-2016-opensubtitles2016,lison-etal-2018-opensubtitles2018},
collected from movie subtitles, contains 1.1M Japanese-Chinese
sentence pairs.


The largest publicly available Japanese-Chinese parallel data
collected from the web is CCMatrix \cite{schwenk-etal-2021-ccmatrix},
with approximately 12M sentence pairs.  JParaCrawl Chinese v1.0
\cite{morishita-etal-2020-jparacrawl} contains 83K Japanese-Chinese
sentence pairs. The Asian Language Treebank
\cite{thu-etal-2016-introducing} translates English Wikinews into
Japanese, Chinese, and other Asian languages and contains
approximately 20,000 sentences divided into train, dev, and test sets.


\section{Methodology}

\subsection{Parallel website mining}

\begin{table*}
  \centering
  \begin{tabular}{lrrrrr}
    \hline
    & \#URLs & \#errors & \#crawled & \#extracted (rate) & \#sentences \\
    \hline
    Common Crawl	& 40,000 & 19,878 & 20,122 & 5,483 (0.272) & 2,786,467 \\
    Crowdsourcing	& 11,184 &    168 & 11,016 & 8,204 (0.745) & 4,602,328 \\
    \hline
  \end{tabular}
  \caption{Comparison of web mining method: Common Crawl vs. crowdsourcing}
  \label{tab:web_mining_methods}
\end{table*}

Our procedure for collecting parallel data is the same as ParaCrawl
\cite{banon-etal-2020-paracrawl} and follows the pipeline of
Bitextor\footnote{\url{https://github.com/bitextor/bitextor}}, which
consists of web crawling, document alignment, sentence alignment, and
parallel corpus filtering.

In ParaCrawl, they determine which websites to crawl by analyzing the
Common Crawl archives. They first apply CLD2
\footnote{\url{https://github.com/CLD2Owners/cld2}} to each web page
to identify its language and extract websites containing approximately
the same target language pair texts. They then crawl the extracted
websites with Heritrix
\footnote{\url{https://github.com/internetarchive/heritrix3}}.

In this study, we analyzed 12 sets of Common Crawl archives (104TB in
total) published from September 2021 to June 2023 using the language
detector CLD2 and enumerated about 40,000 websites that contain
roughly equal amounts of Japanese and Chinese text in order of total
text volume in a website.
We used Extractor\footnote{\url{https://github.com/paracrawl/extractor}} from the ParaCrawl project for this procedure.

\citet{morishita-etal-2022-domain} proposed a method of collecting
parallel URLs (parallel web pages) using cloud workers for domain
adaptation of machine translation.  We asked crowd workers to collect
websites containing parallel pages, specifying only the language
pairs, and report the pair of URLs of the Japanese and Chinese top
pages for each website.

For both parallel top page URL pairs collected using crowdsourcing and
bilingual website URLs obtained from Common Crawl, we used Heritrix to
crawl each website for up to 48 hours, crawling Word and PDF files as
well as HTML.

\subsection{Document and sentence alignment}

For document and sentence alignment in Bitextor, we can use either
machine translation or a bilingual dictionary to determine semantic
equivalence.  This study used bilingual dictionary-based document and
sentence alignment to create a parallel corpus with minimum external
language resources required.

The bilingual dictionary-based document alignment in Bitextor
calculates the similarity of documents from features obtained from the
bilingual dictionary and the structure of HTML.  The bilingual
dictionary-based sentence alignment uses Hunalign
\cite{Varga_etal_RANLP2005}\footnote{\url{http://mokk.bme.hu/resources/hunalign/}}.

We used mecab\footnote{\url{https://taku910.github.io/mecab/}} for
Japanese word segmentation and
jieba\footnote{\url{https://github.com/fxsjy/jieba}} for Chinese word
segmentation.  We used the EDR Japanese-Chinese bilingual Dictionary
(533,957 entries)
\cite{zhang-etal-2007-building}\footnote{\url{https://www2.nict.go.jp/ipp/EDR/JPN/J_HotNews.html}}
as our bilingual dictionary.

To reduce the computation of document and sentence alignment, we
applied word segmenters to Japanese and Chinese headwords in the EDR
Japanese-Chinese Bilingual Dictionary to obtain 157,900 one-to-one
alignment word pairs.  We added correspondences between Japanese Kanji
and simplified Chinese characters to the bilingual dictionary,
resulting in approximately 160,000 entries.
 
\subsection{Parallel corpus filtering}

The ParaCrawl project has two parallel corpus filters, Bicleaner and
Bicleaner AI.  Bicleaner
\cite{sanchez-cartagena-etal-2018-prompsits,ramirez-sanchez-etal-2020-bifixer}
extracts features using word translation probabilities and statistical
language models and trains a random-forest classifier to classify
whether a sentence pair is parallel.  Bicleaner AI
\cite{zaragoza-bernabeu-etal-2022-bicleaner} uses a pre-trained
multilingual language model to train a binary classifier.  Both
methods require high-quality parallel data to train the classifier.

In this study, we used
Bicleaner\footnote{\url{https://github.com/bitextor/bicleaner}} to
minimize the use of external resources and improve computational
efficiency.  We used mecab and pkuseg
\cite{Luo_etal_arXiv2019}\footnote{\url{https://github.com/lancopku/pkuseg-python}}
for Japanese and Chinese word segmentation to compute word frequency
and obtain word alignment from high-quality parallel sentence pairs.
We used AWESOME-align \cite{dou-neubig-2021-word} for word alignment
to compute word translation probabilities.

To train the parallel corpus filter, we used an in-house
Japanese-Chinese parallel corpus (1.2M sentence pairs) consisting of
travel conversations, dictionary examples, literary works, and
newspaper articles.  Among the in-house Japanese-Chinese parallel
corpus, the Basic Travel Expression Corpus (BTEC, about 0.5M sentence
pairs) \cite{takezawa-etal-2002-toward} is the largest, followed by
the dictionary example sentences (about 260,000 sentence pairs).

We used the Japanese-Chinese Bicleaner model to compute scores for
bilingual sentence pairs and extract sentence pairs with a threshold
value of 0.5 or higher.  We further calculated each sentence's vector
using the multilingual sentence embedding model LaBSE
\cite{feng-etal-2022-language} and filtered out sentence pairs with a
cosine distance below a threshold of 0.7.

For URLs obtained from Common Crawl and URL pairs obtained from
crowdsourcing, Table~\ref{tab:web_mining_methods} shows the number of
sites we successfully crawled, the number of sites that yielded any
parallel sentence pairs, and the total number of parallel sentence
pairs obtained.  The percentage of successful parallel sentence pair
extraction was considerably higher for websites obtained from
crowdsourcing, at 74.5 percent, compared to 27.2 percent for those
obtained from Common Crawl.  The total number of parallel sentence
pairs obtained for websites from Common Crawl and those from
crowdsourcing is 2.8M and 4.6M sentence pairs, respectively,
indicating that crowdsourcing can collect more sentence pairs with
less crawling than analyzing Common Crawl.



\section{Translation experiments}

\subsection{Datasets}

\begin{table*}
  \centering
  \begin{tabular}{lrrr}
    \hline
    & train & dev & test  \\
    \hline
    CCMatrix		& 12,403,136 \\
    WikiTitles		&  1,661,273 \\
    WikiMatrix		&  1,325,674 \\
    OpenSubtitles2018	&  1,091,295 \\
    Crowdsourcing (ours)	& 4,643,867 \\
    \hline
    news-commentary-v18	&	& 1,625 \\
    Asian Language Treebank &	& 1,000	& 1,000 \\
    ASPEC-JC		&	&	& 2,107 \\
    FLORES-200		&	&   997	& 1,012 \\
    NTREX-128		&	&	& 1,997 \\
    bitext\_cj		&	&	& 1,000 \\
    WMT2023j		&	&	&   992 \\
    \hline
    total		& 2,1125,245 & 3,622 & 8,126 \\
    \hline
  \end{tabular}
  \caption{Japanese Chinese Parallel Datasets used in the experiments.}
  \label{tab:Parallel_Datasets}
\end{table*}

We examined the accuracy of Japanese-to-Chinese and
Chinese-to-Japanese translations to assess the quality of parallel
sentence pairs collected using crowdsourcing.
Table~\ref{tab:Parallel_Datasets} shows the Japanese-Chinese parallel
datasets used in the translation experiments.

We used CCMatrix \cite{schwenk-etal-2021-ccmatrix},
WikiTitles \cite{tiedemann-2012-parallel},
WikiMatrix \cite{schwenk-etal-2021-wikimatrix}, and
OpenSubtitles2018 \cite{lison-etal-2018-opensubtitles2018} for
comparison because they have more than one million sentence pairs and
are readily available.
We combined WikiTitles, WikiMatrix, and OpenSubtitles2018 into one
(wt-wm-os), and trained three models from ccmatrix, wt-wm-os, and
crowdsourcing.
We further trained one model from all five corpora.

For the development set, we used
news-commentary-v18 (1,677 sentence
pairs)\footnote{\url{https://data.statmt.org/news-commentary/v18.1/}},
the dev set of Asian Language Treebank Parallel Corpus (1,000 sentence
pairs)\footnote{\url{https://www2.nict.go.jp/astrec-att/member/mutiyama/ALT/index.html}},
and the dev set of FLORES-200 (997 sentence
pairs)\footnote{\url{https://github.com/facebookresearch/flores}}.

For the test sets, we used the test set of Asian Language Treebank
(1,000 sentence pairs), the test set of ASPEC-JC (2107 sentence pairs),
the devtest of FLORES-200  (1012 sentence pairs), and NTREX-128 (1997
sentence pairs) \cite{federmann-etal-2022-ntrex}.  We also used as our
test set 1,000 sentences randomly sampled from our in-house
Japanese-Chinese parallel corpus (bitext\_cj) and our in-house Chinese
translations of news (495 sentences) and question answering (497
sentences) from the WMT2023 Japanese-English test set (wmt2023j).  The
source language of these test sets is Japanese for aspec-jc and
wmt2023j, a mixture of Japanese and Chinese for bitext\_cj, and
English for the others.


\subsection{Experiment condition}
\begin{table}[!ht]
  \small
  \centering
  \begin{tabular}{ll}
    \hline
    architecture	& transformer\_wmt\_en\_de\_big \\
    enc-dec layers	& 6 \\
    optimizer		& Adam ($\beta_{1}=0.9, \beta_{2}=0.98$) \\
    learning rate schedule	& inverse square root decay \\
    warmup steps	& 4,000 \\
    max learning rate	& 0.001 \\
    dropout		& 0.3 \\
    gradient clip	& 0.1 \\
    batch size		& 1M tokens \\
    max number of updates	& 60K steps \\
    validate interval updates	& 1K steps \\
    patience		& 5 \\
    \hline
  \end{tabular}
  \caption{Hyper parameters of Transformer}
  \label{tab:parameters}
\end{table}

We used fairseq \cite{ott-etal-2019-fairseq} as the translation
software and transformer big \cite{vaswani-etal-2017-transformer} as the
translation model.
Table~\ref{tab:parameters} shows the hyper parameters of Transformer.
We used sentencepiece \cite{kudo-richardson-2018-sentencepiece} to
tokenize training, development, and test data.
The vocabulary size is 32K for both Japanese and Chinese.
We evaluated translation accuracy using sacreBLEU
\cite{papineni-etal-2002-bleu,post-2018-call} and COMET
(wmt22-comet-da) \cite{rei-etal-2020-comet}.

\subsection{Translation accuracy}
\begin{table*}
  \centering
  \begin{tabular}{lrrrrrr|rr}
\hline
	& \multicolumn{2}{c}{ccmatrix}	& \multicolumn{2}{c}{wt-wm-os}	& \multicolumn{2}{c}{crowdsourcing}	& \multicolumn{2}{|c}{all} \\
test set	&bleu	&comet	&bleu	&comet	&bleu	&comet	&bleu	&comet	\\
\hline
alt\_test	&34.4 	&0.856 	&18.9 	&0.779 	&35.8 	&0.847 	&38.0 	&0.867 \\
aspec\_test	&35.8 	&0.856 	&17.6 	&0.767 	&37.8 	&0.862 	&37.5 	&0.863 \\
flores200\_devtest	&29.5 	&0.860 	&16.0 	&0.776 	&33.8 	&0.863 	&35.1 	&0.872 \\
ntrex128	&25.2 	&0.815 	&14.2 	&0.735 	&25.4 	&0.806 	&27.5 	&0.823 \\
bitext\_cj\_test	&22.3 	&0.808 	&11.9 	&0.739 	&23.8 	&0.812 	&24.9 	&0.825 \\
wmt2023j	&23.9 	&0.801 	&11.7 	&0.713 	&32.1 	&0.824 	&29.5 	&0.826 \\
\hline
average	&28.5 	&0.833 	&15.1 	&0.751 	&\textbf{31.5} 	&\textbf{0.836} 	&32.1 	&0.846 \\
\hline
  \end{tabular}
  \caption{Japanese-to-Chinese translation}
  \label{tab:J2C}
\end{table*}

\begin{table*}
  \centering
  \begin{tabular}{lrrrrrr|rr}
\hline
	& \multicolumn{2}{c}{ccmatrix}	& \multicolumn{2}{c}{wt-wm-os}	& \multicolumn{2}{c}{crowdsourcing}	& \multicolumn{2}{|c}{all} \\
testiest	&bleu	&comet	&bleu	&comet	&bleu	&comet	&bleu	&comet	\\
\hline
alt\_test	&24.1	&0.886 	&15.9	&0.817 	&22.6 	&0.872 	&25.0 	&0.890 \\
aspec\_test	&29.7	&0.896 	&19.9	&0.834 	&29.9 	&0.897 	&31.7 	&0.902 \\
flores200\_devtest	&26.2	&0.887 	&14.4	&0.795 	&26.6 	&0.881 	&27.9 	&0.893 \\
ntrex128	&18.8	&0.859 	&11.8	&0.775 	&17.5 	&0.844 	&19.8 	&0.865 \\
bitext\_cj\_test	&17.6	&0.833 	&8.8	&0.755 	&16.5 	&0.833 	&18.2 	&0.847 \\
wmt2023j	&24.9	&0.874 	&14.1	&0.782 	&23.6 	&0.878 	&26.9 	&0.889 \\
\hline
average	&\textbf{23.55}	&\textbf{0.872} 	&14.15	&0.793 	&22.8 	&0.867 	&24.9 	&0.881 \\
\hline
  \end{tabular}
  \caption{Chinese-to-Japanese translation}
  \label{tab:C2J}
\end{table*}

Table~\ref{tab:J2C} shows the translation accuracy from Japanese to
Chinese, and Table~\ref{tab:C2J} shows that from Chinese to Japanese.

Among the three translation models, ccmatrix, wt-wm-os, and
crowdsourcing, ccmatrix and crowdsourcing have about the same translation
accuracy, while wt-wm-os is less accurate.
Between ccmatrix and crowdsourcing, crowdsourcing has higher
translation accuracy from Japanese to Chinese, and ccmatrix has higher
accuracy from Chinese to Japanese.
Creating a single translation model from all bilingual data yields a
higher translation accuracy than these three models.

\section{Discussion}
Crowdsourcing (4.6M) has only one-third of sentence pairs of CCMatrix
(12.4M), but the translation accuracy is about the same.  This
indicates that parallel sentence pairs obtained using crowdsourcing
have higher quality than CCMatrix.

For Japanese-to-Chinese translation, the higher accuracy of our
parallel corpus collected using crowdsourcing compared to CCMatrix is
probably due to the fact that the crowdsourcing was done in Japan by
Japanese crowd workers.  Many of the websites collected by Japanese
crowd workers are Japanese websites that include pages translated into
Chinese.  More diversity may be needed when translating Chinese into
Japanese.

This study evaluated the translation accuracy of parallel sentence
pairs collected using crowdsourcing (4.6M).  However, we expect that
adding parallel sentence pairs collected using Common Crawl (2.8M)
will increase diversity and improve translation accuracy from Chinese
to Japanese. Another issue is that evaluating Chinese-to-Japanese
translation accuracy could be more reliable if we had a test set whose
source sentences originated from Chinese and whose reference sentences
are direct manual translations from Chinese to Japanese.

\section{Conclusion}
This paper describes an attempt to create a Japanese-Chinese parallel
corpus from the web by collecting URL pairs of parallel websites
through crowdsourcing.  We collected 4.6M sentence pairs and showed
that we could achieve the same level of translation accuracy as the
CCMatrix (12.4M) with one-third of the data.

In the future, we will create an adult filter to filter the parallel
sentence pairs (2.8M) collected using Common Crawl and add these to
make a Japanese-Chinese parallel corpus with more diverse content.  We
will also train a machine translation model using the sentence pairs
created with bilingual dictionary-based document and sentence
alignment to perform machine translation-based document and sentence
alignment, which could improve the quality of parallel sentence pairs.

\bibliography{anthology,custom}




\end{document}